\documentclass[10pt,twocolumn,letterpaper]{article}

\usepackage{cvpr}              
\usepackage{dsfont}
\usepackage{makecell}
\usepackage{multirow}
\usepackage{booktabs}
\usepackage{siunitx}
\usepackage{graphicx}
\definecolor{cvprblue}{rgb}{0.21,0.49,0.74}
\usepackage[pagebackref,breaklinks,colorlinks,allcolors=cvprblue]{hyperref}

\def\paperID{44914} 
\def\confName{CVPR}
\def\confYear{2026}

\title{MV3DIS: Multi-View Mask Matching via 3D Guides for Zero-Shot 3D~Instance~Segmentation}

\author{
    Yibo Zhao$^{1}$ \quad 
    Yigong Zhang$^{1}$ \quad 
    Jin Xie$^{2}$\thanks{Corresponding author.} \\
    $^1$College of Computer Science, Nankai University \\
    $^2$School of Intelligence Science and Technology, Nanjing University \\
    {\tt\small yibozhao@mail.nankai.edu.cn, zyg025@nankai.edu.cn, csjxie@nju.edu.cn}
}
\begin{document}
\maketitle
\begin{abstract}
Conventional 3D instance segmentation methods rely on labor-intensive 3D annotations for supervised training, which limits their scalability and generalization to novel objects. Recent approaches leverage multi-view 2D masks from the Segment Anything Model (SAM) to guide the merging of 3D geometric primitives, thereby enabling zero-shot 3D instance segmentation. However, these methods typically process each frame independently and rely solely on 2D metrics, such as SAM prediction scores, to produce segmentation maps. This design overlooks multi-view correlations and inherent 3D priors, leading to inconsistent 2D masks across views and ultimately fragmented 3D segmentation. In this paper, we propose MV3DIS, a coarse-to-fine framework for zero-shot 3D instance segmentation that explicitly incorporates 3D priors. Specifically, we introduce a 3D-guided mask matching strategy that uses coarse 3D segments as a common reference to match 2D masks across views and consolidates multi-view mask consistency via 3D coverage distributions. Guided by these view-consistent 2D masks, the coarse 3D segments are further refined into precise 3D instances. Additionally, we introduce a depth consistency weighting scheme that quantifies projection reliability to suppress ambiguities from inter-object occlusions, thereby improving the robustness of 3D-to-2D correspondence. Extensive experiments on the ScanNetV2, ScanNet200, ScanNet++, Replica, and Matterport3D datasets demonstrate the effectiveness of MV3DIS, which achieves superior performance over previous methods. Code is available at \href{https://github.com/zybjn/MV3DIS}{https://github.com/zybjn/MV3DIS}.
\end{abstract}
\section{Introduction}
\begin{figure}[t!]
    \centering
    \includegraphics[width=\columnwidth]{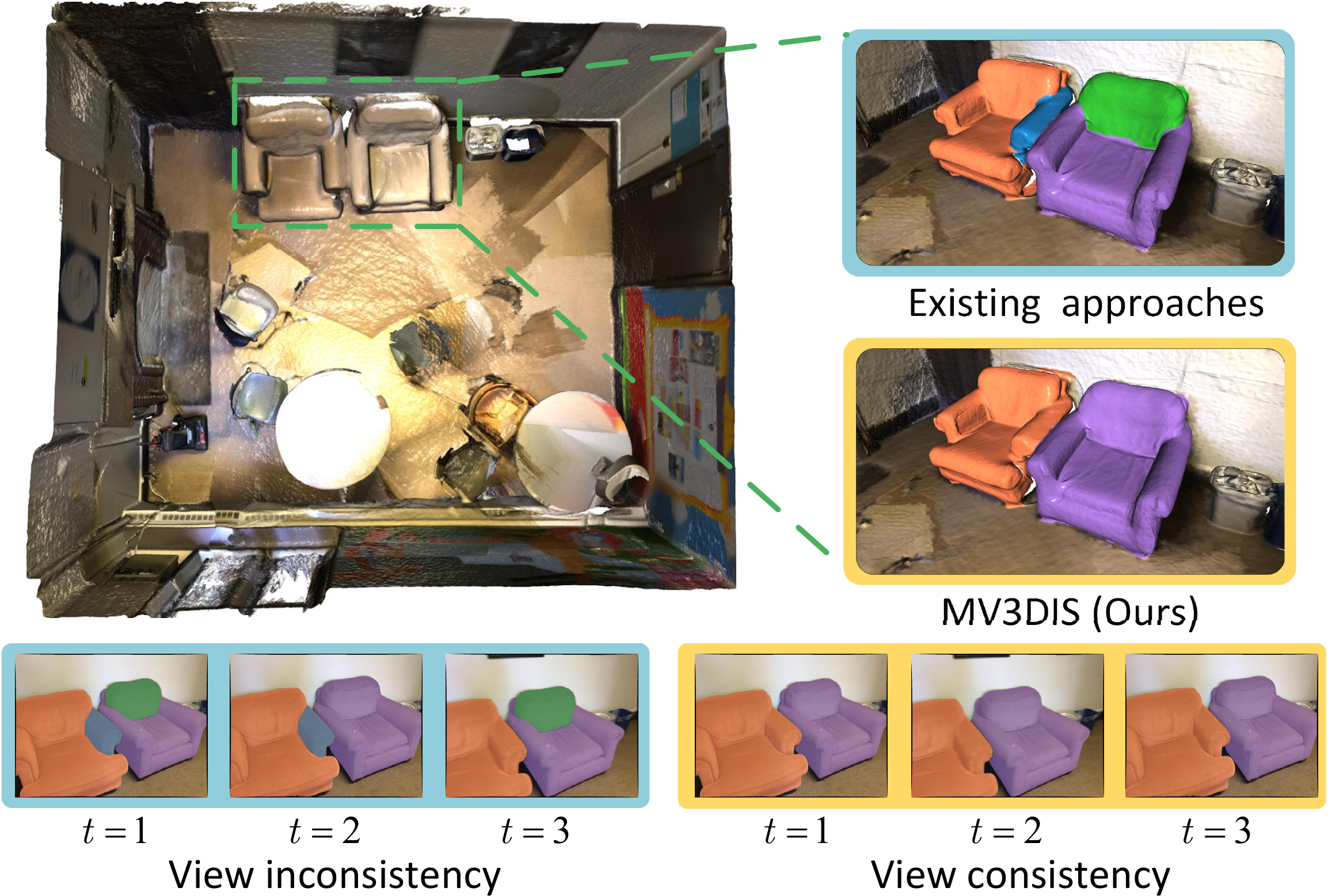}
    \caption{\textbf{View inconsistency in existing methods versus view consistency with MV3DIS.}
Top-right: 3D instance segmentations from existing approaches and MV3DIS.
Bottom blue and yellow strips: 2D segmentation maps from different views produced by existing approaches and MV3DIS.
Existing methods yield view-inconsistent masks and fragmented 3D instances, whereas MV3DIS enforces view consistency and produces coherent 3D instance segmentations.}
    \label{introduce}
\end{figure}

3D instance segmentation is a fundamental task in 3D scene understanding and has been widely applied in areas such as autonomous driving and robotic navigation.
Traditional 3D instance segmentation methods \cite{mask3d,vu2022softgroup} depend on supervised training with annotated 3D data, achieving strong performance in a closed-set setting.
However, the high cost of collecting and annotating 3D data presents significant challenges in building large-scale, high-quality datasets, limiting model generalization to previously unseen objects.
In contrast, the lower cost and ease of annotating 2D images have facilitated the creation of large-scale 2D image datasets. 
Visual foundation models \cite{clip,sam,groundingdino} trained on these internet-scale datasets exhibit remarkable capabilities.
Recently, the Segment Anything Model (SAM) \cite{sam} has demonstrated impressive zero-shot segmentation performance.
Inspired by this, several studies \cite{sam3d,sai3d,open3dis} have explored lifting 2D segmentation masks to 3D instance segmentation, thereby alleviating reliance on annotated 3D data.
Specifically, these methods leverage per-frame 2D segmentation masks from SAM together with 3D geometric properties to merge 3D points into instance-level segments.

However, existing methods often exhibit poor cross-view consistency in 2D segmentation maps, leading to over-segmentation in 3D scenes.
Typically, these methods process each frame independently and rely solely on 2D metrics (e.g., SAM prediction scores) to produce segmentation maps, while neglecting the inherent 3D priors.
For instance, SAM3D \cite{sam3d} and SAI3D \cite{sai3d} perform automatic mask segmentation via dense grid point prompts on the image plane and use 2D results to guide 3D primitive merging.
To improve efficiency, these methods typically merge multiple independent masks within each frame into a single segmentation map.
However, as shown in \cref{introduce}, this process may cause fragmented 2D masks to overwrite complete instance masks, leading to cross-view inconsistencies for the 2D segmentations of the same 3D object.
These inconsistencies ultimately lead to fragmented 3D segmentations.
Furthermore, SAM2Object \cite{sam2object} enhances view consistency using SAM2 tracking capabilities. However, such tracking-based methods are susceptible to video quality issues, such as motion blur, and may lead to error propagation. Moreover, they are limited to continuous video sequences and struggle to handle sparse multi-view datasets.

To address these limitations, we introduce MV3DIS, a novel zero-shot 3D instance segmentation method that leverages SAM's powerful segmentation capabilities.
MV3DIS employs a coarse-to-fine pipeline to enhance multi-view 2D mask consistency by integrating 3D information, thereby improving 3D instance segmentation.
Specifically, we propose a 3D-guided mask matching strategy that segments the point cloud into initial 3D segments and uses their projections as common references to match consistent 2D masks across views.
These projections provide rich 3D priors, significantly enhancing multi-view mask consistency.
The resulting consistent 2D masks further refine the initial 3D segmentation, yielding fine-grained instance results.
Furthermore, we observe that inter-object occlusions in projecting 3D points can lead to erroneous matching between 3D segments and 2D masks, thereby affecting subsequent mask matching and 3D segmentation.
To mitigate this, we propose a depth consistency weight that quantifies projection reliability to suppress ambiguities caused by occlusions, thus enhancing the robustness of 3D-to-2D correspondence.

Overall, our contributions are summarized as follows:
\begin{itemize}
    \item We propose MV3DIS, a zero-shot 3D instance segmentation framework that leverages SAM. It enforces multi-view consistency through 3D-guided mask matching, thereby improving 3D segmentation quality.
	\item We propose a depth consistency weight that evaluates projection reliability to establish more accurate 2D-to-3D correspondences, thereby improving 3D segmentation robustness.
    \item Extensive experiments on the ScanNetV2, ScanNet200, ScanNet++, Replica, and Matterport3D datasets demonstrate the effectiveness of our method.
\end{itemize}

\section{Related work}
\textbf{Closed-vocabulary 3D segmentation.}
The task of 3D instance segmentation aims to identify foreground objects in a scene and assign a unique label to each instance.
Existing approaches for 3D instance segmentation can be classified into proposal-based, grouping-based, and transformer-based methods.
Proposal-based methods typically predict preliminary candidate regions, which are then refined for segmentation and classification \cite{yang2019learning,yi2019gspn,hou20193d,engelmann20203d}. 
Grouping-based methods directly extract features from point cloud data, grouping adjacent points into object instances through clustering or similarity measures \cite{jiang2020pointgroup,liang2021instance,chen2021hierarchical,vu2022softgroup}.
In contrast, transformer-based methods leverage self-attention mechanisms to capture long-range dependencies between points in the point cloud, significantly improving segmentation performance ~\cite{mask3d,sun2023superpoint,lai2023mask,lu2023query,kolodiazhnyi2024oneformer3d}.
Among these, Mask3D~\cite{mask3d} introduces the first Transformer-based method for 3D instance segmentation, leveraging a transformer decoder to directly predict instance masks from 3D point clouds, achieving state-of-the-art performance.
Despite notable progress, these methods still heavily rely on large amounts of annotated 3D data for training, which limits their generalization ability in open-world scenarios.

\noindent\textbf{Open-vocabulary 2D Recognition.}
Open-vocabulary 2D recognition aims to enable models to identify novel object categories unseen during training. Depending on the task type, these can be categorized as open-vocabulary object detection (OVOD) \cite{zhong2022regionclip, wang2023object, yao2023detclipv2, zang2022open, kaul2023multi, pham2024lp, groundingdino, yoloworld}, open-vocabulary semantic segmentation (OVSS) \cite{bucher2019zero, li2022language, ghiasi2022scaling, xu2022groupvit, ding2022decoupling, liang2023open, li2023open, wu2023diffumask, zou2023segment}, and open-vocabulary instance segmentation (OVIS) \cite{huynh2022open,vs2023mask,wu2023betrayed,zhang2023simple,li2023semantic,samhq}.
Recently, SAM \cite{sam} made significant advances in 2D segmentation by enabling the segmentation of any object in an image through prompts.
SAM2 \cite{sam2} further extended SAM to video segmentation by storing information about the object and previous interactions, enabling the refinement of mask predictions across frames.
In this work, we leverage the strong zero-shot generalization ability of SAM for 3D segmentation, thereby alleviating the need for manual 3D annotations and training.

\noindent\textbf{Open-Vocabulary 3D segmentation.}
Inspired by the generalization capabilities of foundation models \cite{sam,sam2,yoloworld,groundingdino,clip}, several methods \cite{hegde2023clip,jatavallabhula2023conceptfusion,kerr2023lerf,lee2024segment,huang2024openins3d,ding2023pla,openscene,li2025ovseg3r,
tang2025onlineanyseg} have leveraged the powerful zero-shot recognition abilities of 2D vision-language models to recognize previously unseen categories in 3D scenes.
OpenMask3D \cite{openmask3d} and OpenYOLO3D \cite{openyolo3d} utilize the 3D instance segmenter Mask3D \cite{mask3d} to extract category-agnostic 3D instances and combine 2D foundation models \cite{sam,yoloworld,clip} for open-vocabulary queries.
However, since pre-trained 3D models struggle to segment uncommon objects, these methods face challenges in recognizing tail classes.
Recently, several studies \cite{sam3d,sai3d,open3dis,sampro3d,sam2object,any3dis,jung2025details,ovir3d,segment3d,maskclustering,wang2025sgs} have explored leveraging SAM to segment multi-view images and lifting the results to 3D space. 
SAM3D \cite{sam3d} pioneered the use of 2D segmentation masks from SAM to iteratively merge partial 3D point cloud masks, ultimately achieving 3D segmentation of the entire scene.
Methods such as SAI3D \cite{sai3d}, Open3DIS \cite{open3dis}, and SAM-graph \cite{samgraph} represent the scene as a superpoint graph and employ node merging algorithms guided by multi-view segmentation masks to aggregate 3D superpoints into instances.
However, these methods often suffer from multi-view segmentation inconsistencies, leading to fragmented 3D segmentation results.
SAM2Object \cite{sam2object} enhances view consistency by leveraging SAM2 powerful tracking capabilities. However, such tracking-based approaches are susceptible to video quality issues and struggle with sparse multi-view datasets.

\section{Method}
\begin{figure*}[t]
    \centering
    \includegraphics[width=0.99\textwidth]{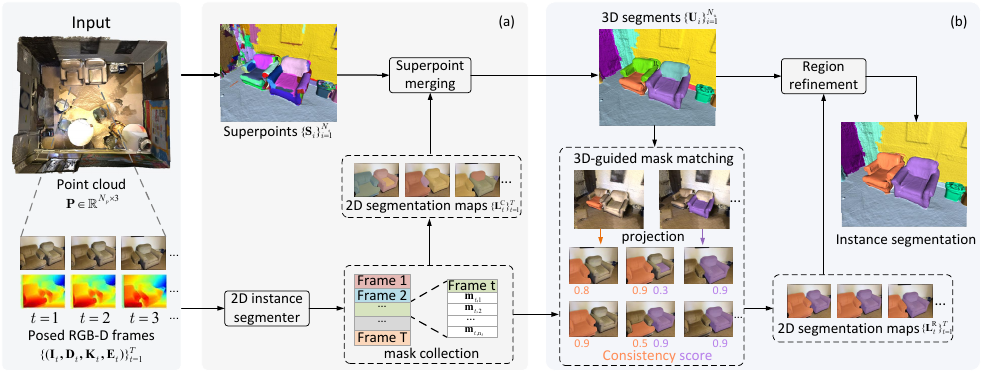}
    \caption{\textbf{Method overview.}
Our method consists of two stages: (a) Coarse 3D Instance Segmentation and (b) 3D Instance Refinement.
We first over-segment the point cloud into superpoints and use SAM to generate per-frame 2D masks, which are organized into a global mask collection.
These masks are consolidated into coarse 2D segmentation maps that guide the merging of superpoints into coarse 3D segments.
We then project the coarse 3D segments onto images and perform 3D-guided mask matching to associate consistent 2D masks from the collection.
Finally, we refine the coarse 3D segments into coherent 3D instances using the consistent 2D segmentation maps.
}
    \label{method}
\end{figure*}
\subsection{Overview}
Given a 3D point cloud $\mathbf{P} \in \mathbb{R}^{N_p \times 3}$ and a sequence of $T$ RGB-D frames
$\left\{ (\mathbf{I}_t, \mathbf{D}_t, \mathbf{K}_t, \mathbf{E}_t) \right\}_{t=1}^{T}$, where $\mathbf{I}_t$, $\mathbf{D}_t$, $\mathbf{K}_t$, and $\mathbf{E}_t$ denote the RGB image, depth image, camera intrinsics, and extrinsics of the $t$-th frame, respectively.
Our objective is to estimate instance masks for all foreground objects in the scene.
As illustrated in \cref{method}, our coarse-to-fine framework begins by over-segmenting the point cloud into superpoints $\{\mathbf{S}_i\}_{i=1}^{N_s}$ using a graph-cut algorithm~\cite{superpoint}.
In the coarse 3D instance segmentation stage (\cref{Coarse3DIS}), we derive 2D segmentation maps from SAM predictions to guide superpoint merging into coarse 3D segments $\{\mathbf{U}_i\}_{i=1}^{N_u}$.
Next, in the 3D instance refinement stage (\cref{3DIR}), we introduce a 3D-guided mask matching method to match consistent 2D masks across views, which are then used to refine the coarse 3D segments into coherent instance segmentations.
In \cref{2D3DSC}, we detail the superpoint merging and region refinement algorithms.

\subsection{Superpoints and SAM-Based Coarse 2D Maps}\label{Coarse3DIS}
The coarse 3D instance segmentation stage, illustrated in \cref{method}~(a), consists of three main components: superpoint generation, construction of coarse 2D segmentation maps, and superpoint merging (\cref{2D3DSC}).

\noindent\textbf{Superpoints.}
In the preprocessing step, we use the graph-cut algorithm \cite{superpoint} to over-segment the point cloud into superpoints.
Similar to superpixels in 2D images, superpoints are a set of 3D points that are compactly distributed in 3D space with similar geometric properties. The number of superpoints is much smaller than the original points, which not only enhances the efficiency of our framework but also effectively incorporates geometric priors.

\noindent\textbf{Coarse 2D segmentation maps.}
We employ 2D foundation models, such as Grounding-DINO~\cite{groundingdino} and SAM~\cite{sam}, to segment per-frame 2D instance masks.
Following previous works \cite{sam3d}, we apply NMS based on the SAM prediction IoU scores to remove redundant masks and merge the remaining masks in each frame into a single coarse segmentation map $\{\mathbf{L}^\mathrm{C}_t\}_{t=1}^{T}$.
In the segmentation map, each pixel is assigned a mask label indicating the index of the covering 2D mask. 
When a pixel is covered by multiple masks, only the one with the highest prediction IoU score is retained, ensuring non-overlapping masks.
However, coarse segmentation maps often lack multi-view consistency, and 2D-only post-processing can cause fragmented masks to overwrite complete instance masks, leading to fragmented 3D segmentation~\cite{sam3d,sai3d}.

\subsection{3D-Guided Multi-View Mask Matching}\label{3DIR}
We propose a 3D-guided mask matching method that enhances multi-view mask consistency for refined 3D segmentation.
The core of our method is to leverage coarse 3D segment projections as a common reference to associate highly correlated 2D masks and evaluate their consistency based on the 3D coverage distribution.

\noindent\textbf{Point cloud projection.}
We project the 3D point cloud $\mathbf{P}$ onto the $t$-th frame using a pinhole camera model with intrinsics $\mathbf{K}_t$ and extrinsics $\mathbf{E}_t$.
For each 3D point in homogeneous form \mbox{$\mathbf{p} = (x, y, z, 1)^{\top}$},
we compute its camera coordinates as
$\mathbf{p}_c = (x_c, y_c, z_c)^{\top} = \mathbf{E}_t \mathbf{p}$.
Its homogeneous image coordinates are
$\tilde{\mathbf{p}} = (\tilde{u}, \tilde{v}, \tilde{w})^{\top} = \mathbf{K}_t \mathbf{p}_c$,
and the corresponding pixel location is
$(u, v) = (\tilde{u} / \tilde{w},\, \tilde{v} / \tilde{w})$.
A point is considered visible if it lies inside the image and satisfies the depth consistency condition.
We define a binary visibility indicator $\mathcal{I}_{\text{vis}}(\mathbf{p}, \mathbf{I}_t)$ for point $\mathbf{p}$ in the $t$-th frame:
\begin{equation}
\label{pt}
\begin{split}
    \mathcal{I}_{\text{vis}}(\mathbf{p}, \mathbf{I}_t)
    =\,& \mathds{1}(0 \le u < W \wedge 0 \le v < H) \\
       &\cdot \mathds{1}(|z_c - d| < \alpha d),
\end{split}
\end{equation}
where $\mathds{1}(\cdot)$ denotes the indicator function, 
$W$ and $H$ are the image width and height, 
$z_c$ is the projected depth, 
$d$ is the measured depth from the depth image at pixel $(u, v)$, 
and $\alpha = 0.05$ is a relative tolerance threshold that improves robustness to varying distances from the camera.

\begin{figure}[t]
    \centering
    \includegraphics[width=0.97\columnwidth]{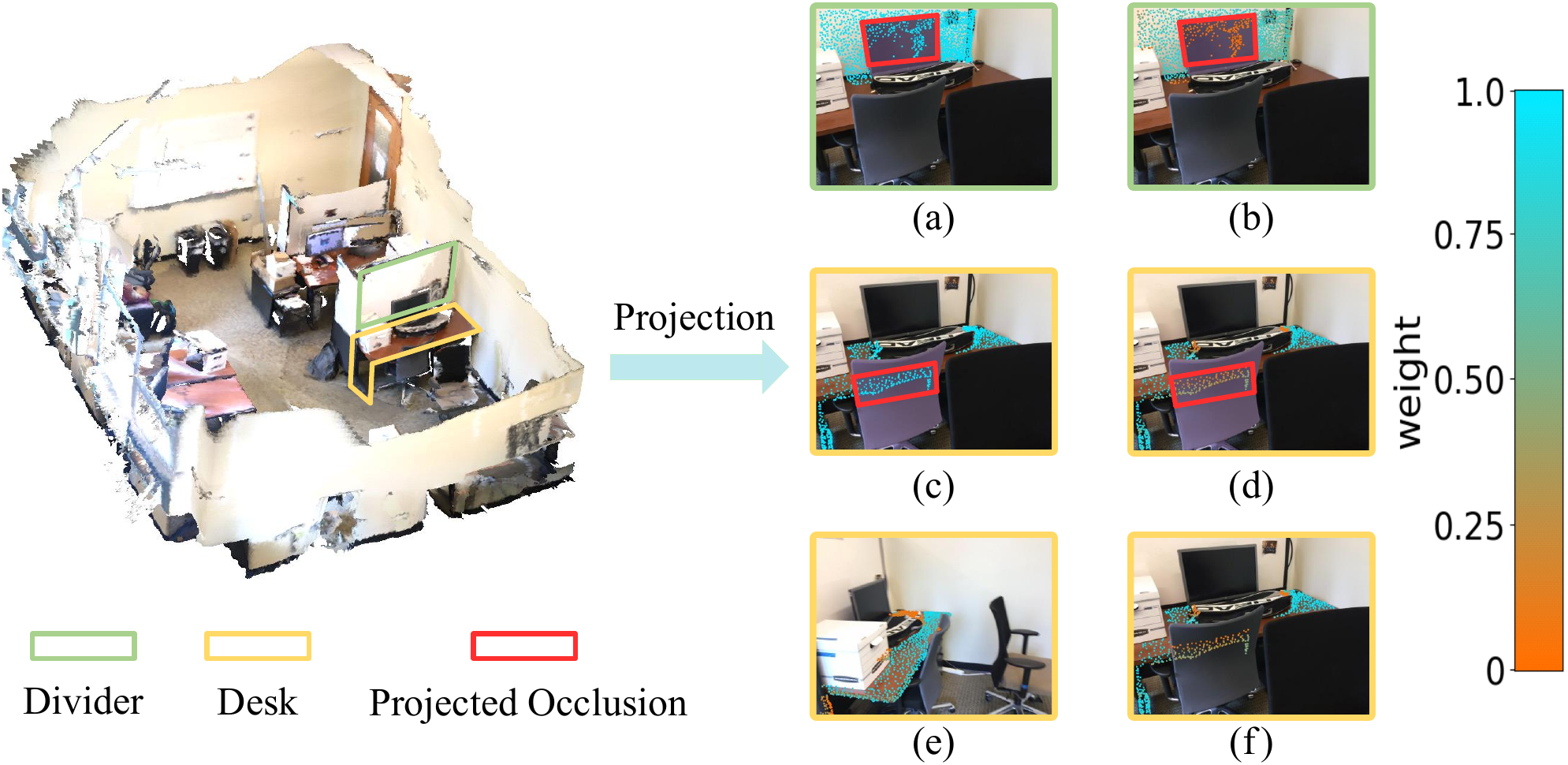}
    \caption{
\textbf{Depth Consistency and Visibility Weights.} During 3D-to-2D projection, near-threshold occluded points (red boxes) may be misidentified as visible on foreground objects (a, c).
To mitigate such projection errors, our method assigns a continuous depth consistency weight to each point (b, d), which decreases with larger depth discrepancies.
Additionally, the visibility weight quantifies an object's observability, assigning higher weights to frames with more visible points (e) than to frames affected by occlusion (f).
}
\label{weight} 
\end{figure}

\noindent\textbf{Point depth consistency weight.}
Although \cref{pt} effectively filters points with large depth discrepancies, occluded points near the threshold may still be misidentified as visible.
For example, in Figs.~\ref{weight}~(a) and (c), most occluded points in the projected divider and desk are filtered out, but some are still erroneously retained.
To quantify the reliability of the projected points, we assign a depth consistency weight $w^\mathrm{d}_{\mathbf{p}}$ to each point $\mathbf{p}$:
\begin{equation}
\label{eq:depth_weight}
w^\mathrm{d}_\mathbf{p}= 1-\frac{\lvert z_c - d \rvert}{\alpha d}
\end{equation}
As shown in Figs.~\ref{weight}~(b) and (d), larger errors yield lower weights, thus reducing the influence of near-threshold points in subsequent segmentation steps.

\noindent\textbf{Consistent mask selection.}
We enforce multi-view consistency of masks belonging to the same object via a 3D-guided mask matching strategy.
Specifically, as illustrated in \cref{method}~(b), we first project each coarse 3D segment $\mathbf{U}_i$ onto frames to serve as a common reference.
We then match a set of highly correlated candidate masks based on frame visibility and mask visibility.
The frame visibility $\mathcal{V}_{i,t}^\mathrm{f}$ is the ratio of visible projected points from $\mathbf{U}_i$ in the $t$-th frame to its total point count.
Likewise, for $\mathbf{U}_i$, mask visibility $\mathcal{V}^\mathrm{m}_{i,t,j}$ is the ratio of its visible points falling within 2D mask $\mathbf{m}_{t,j}$ to its total visible points in the $t$-th frame.
For each coarse 3D segment $\mathbf{U}_i$, we define the candidate mask set $\mathcal{G}_i$ as:
\begin{equation}
\mathcal{G}_i = \{ (t,j) \mid \mathcal{V}_{i,t}^\mathrm{f} > \tau_\mathrm{f} \text{ and } \mathcal{V}^\mathrm{m}_{i,t,j} > \tau_\mathrm{m} \}
\end{equation}
where $\tau_\mathrm{f} = 0.3$ and $\tau_\mathrm{m} = 0.9$ are predefined thresholds.

View consistent masks are expected to exhibit similar 3D coverage.
To quantify the consistency of matched masks, we define the 3D coverage distribution as the projection occupancy of coarse 3D segments within the 2D mask.
Specifically, for each candidate 2D mask $\mathbf{m}_{t,j}$, we construct a coverage vector $\mathbf{v}_{t,j} \in \mathbb{R}^{N_u}$ over coarse 3D segments, where $N_u$ is the number of coarse 3D segments and the $k$-th entry is given by $\mathbf{v}_{t,j}^k = \bar{w}^\mathrm{d}_{k,t,j} \cdot \mathcal{V}^\mathrm{m}_{k,t,j}$.
Here, the weight $\bar{w}^\mathrm{d}_{k,t,j}\in(0,1)$ measures the average depth consistency of visible points from 3D segment $\mathbf{U}_k$ projected onto $\mathbf{m}_{t,j}$.
For projection errors caused by occlusions, such as points from the divider and desk projected onto the monitor and chair masks in Figs.~\ref{weight}~(b) and (d), the corresponding weights are significantly reduced.
It is defined as:
\begin{equation}
\label{eq:dw}
\bar{w}^\mathrm{d}_{k,t,j} = \frac{\sum_{\mathbf{p} \in \mathbf{U}_k} \mathcal{I}_{\text{vis}}(\mathbf{p},\mathbf{I}_t) \cdot \mathds{1}(\mathbf{p} \in \mathbf{m}_{t,j}) \cdot w^\mathrm{d}_{\mathbf{p}}}{\sum_{\mathbf{p} \in \mathbf{U}_k} \mathcal{I}_{\text{vis}}(\mathbf{p},\mathbf{I}_t) \cdot \mathds{1}(\mathbf{p} \in \mathbf{m}_{t,j})}
\end{equation}
where $\mathcal{I}_{\text{vis}}(\mathbf{p},\mathbf{I}_t)$ is given by \cref{pt}, $w^\mathrm{d}_{\mathbf{p}}$ is defined in \cref{eq:depth_weight}, and $\mathds{1}(\mathbf{p} \in \mathbf{m}_{t,j})$ indicates if the projection of $\mathbf{p}$ lies within the 2D mask $\mathbf{m}_{t,j}$.

Next, for each mask $\mathbf{m}_{t,j}$ in the set $\mathcal{G}_i$, we compute its consistency score $s_{i,t,j}$ as the average cosine similarity between $\mathbf{v}_{t,j}$ and other vectors in $\mathcal{G}_i$:
\begin{equation}
s_{i,t,j} = \frac{1}{|\mathcal{G}_i| - 1} \sum_{(t',j') \in \mathcal{G}_i \setminus \{(t,j)\}} \frac{\mathbf{v}_{t,j} \cdot \mathbf{v}_{t',j'}}{\|\mathbf{v}_{t,j}\| \|\mathbf{v}_{t',j'}\|},
\end{equation}
Since a mask $\mathbf{m}_{t,j}$ may be matched by multiple 3D segments, we take the mean of its scores across different segments as the final consistency score.
We then apply NMS based on the consistency scores to suppress inconsistent masks.
The remaining masks in each frame are merged into a refined segmentation map $\{\mathbf{L}^\mathrm{R}_t\}_{t=1}^{T}$, retaining only the mask with the highest score for overlapping pixels.
Overall, this 3D-guided strategy enforces multi-view mask consistency using 3D geometry, without relying on video tracking or inter-frame temporal coherence.

\subsection{Affinity-Based Merging and Refinement}\label{2D3DSC}
Given the superpoints $\{\mathbf{S}_i\}_{i=1}^{N_s}$ and 2D segmentation maps $\{\mathbf{L}^\mathrm{C}_t\}_{t=1}^{T}$, we construct a graph whose nodes are superpoints, and edges are defined by affinity scores computed from the segmentation maps.
We then employ a region growing algorithm to merge superpoints into coarse 3D segments $\{\mathbf{U}_i\}_{i=1}^{N_u}$.
Finally, using the consistent 2D segmentation maps $\{\mathbf{L}^\mathrm{R}_t\}_{t=1}^{T}$ from mask matching, we refine the assignment of superpoints to coarse segments to obtain fine-grained 3D instance segmentations.

\noindent\textbf{Scene graph construction.}
We define the affinity score between superpoints based on their co-occurrence in the 2D segmentation maps.
For each superpoint $\mathbf{S}_i$, we construct a histogram vector $\mathbf{e}_{i,t}$ over 2D mask labels in the $t$-th segmentation map, where each entry counts the number of visible projected points of $\mathbf{S}_i$ assigned to the corresponding label.
The affinity score $A_{i,j}^t$ between superpoints $\mathbf{S}_i$ and $\mathbf{S}_j$ in the $t$-th frame is computed as the cosine similarity between their histogram vectors:
\begin{equation}
A_{i,j}^t = \frac{\mathbf{e}_{i,t} \cdot \mathbf{e}_{j,t}}{\|\mathbf{e}_{i,t}\| \cdot \|\mathbf{e}_{j,t}\|}
\end{equation}
Due to factors such as projection errors, occlusions, and camera motion, superpoint projection quality and visibility vary across frames.
For example, as shown in Figs.~\ref{weight}~(e) and (f), the desk varies in visibility and occlusion across views.
This implies that the affinity scores computed from each frame are not equally reliable.
Therefore, we compute the overall affinity between superpoints by weighting the affinity scores across all frames using depth consistency and visibility weights.
Specifically, we define the depth consistency weight $\bar{w}^\mathrm{d}_{i,t}\in(0,1)$ for superpoint $\mathbf{S}_i$ in the $t$-th frame as follows:
\begin{equation}
\bar{w}^\mathrm{d}_{i,t} = \frac{\sum_{\mathbf{p} \in \mathbf{S}_i} \mathcal{I}_{\text{vis}}(\mathbf{p}, \mathbf{I}_t) \cdot w^\mathrm{d}_{\mathbf{p}}}{\sum_{\mathbf{p} \in \mathbf{S}_i} \mathcal{I}_{\text{vis}}(\mathbf{p}, \mathbf{I}_t)}
\end{equation}
where $\mathcal{I}_{\text{vis}}(\mathbf{p}, \mathbf{I}_t)$ is given by \cref{pt} and $w^\mathrm{d}_{\mathbf{p}}$ is defined in \cref{eq:depth_weight}.
The visibility weight $\bar{w}^\mathrm{v}_{i,t}\in(0,1)$ for superpoint $\mathbf{S}_i$ is defined as the ratio of its visible points in the $t$-th frame to its total point count.
Thus, the edge weight $\varphi_{i,j,t}$ between superpoints $\mathbf{S}_i$ and $\mathbf{S}_j$ in the $t$-th frame is calculated as:
\begin{equation}
\varphi_{i,j,t} = \bar{w}^\mathrm{d}_{i,t} \, \bar{w}^\mathrm{d}_{j,t} \, \bar{w}^\mathrm{v}_{i,t} \, \bar{w}^\mathrm{v}_{j,t}
\end{equation}
We represent the final affinity matrix between superpoints as $\mathbf{A}\in\mathbb{R}^{N_s \times N_s}$, where the weighted affinity score between $\mathbf{S}_i$ and $\mathbf{S}_j$ is computed as follows:
\begin{equation}
A_{i,j} = 
\frac{
\sum_{t=1}^{T} \varphi_{i,j,t}\,A_{i,j}^t
}{
\sum_{t=1}^{T} \varphi_{i,j,t}
}
\end{equation}

\noindent\textbf{Superpoint merging.}
Based on the constructed scene graph, we employ a region growing algorithm to merge adjacent superpoints into coherent regions.
In region growing, an unassigned superpoint $\mathbf{S}_i$ is merged into region $\mathbf{U}_k$ if it has a neighbor $\mathbf{S}_j \in \mathbf{U}_k$ whose affinity $A_{i,j}$ exceeding the merging threshold $\tau_{\mathrm{merge}}$.
However, this criterion considers only the local pairwise affinity and overlooks the global relationship between $\mathbf{S}_i$ and $\mathbf{U}_k$.
In 3D space, closer superpoints are more likely to belong to the same region, and superpoints with more points provide more representative geometric information.
Inspired by recent works \cite{sai3d,sam2object}, we compute a weighted affinity score by combining the Euclidean distance and point count of each neighbor in $\mathbf{U}_k$, assigning greater weights to closer neighbors with higher point counts.
The merging process iterates until all superpoints are assigned, resulting in the 3D segments~$\{\mathbf{U}_i\}_{i=1}^{N_u}$.

\noindent\textbf{Region refinement.}
In region growing algorithms, superpoints are typically assigned to the first region meeting the merging threshold, rather than the region with the highest affinity score.
These misassignments usually occur at boundary superpoints and can accumulate, adversely affecting the final 3D instance segmentation.
Therefore, we refine the superpoint assignments among the coarse segments using affinity scores recalculated from the refined 2D segmentation maps $\{\mathbf{L}^\mathrm{R}_t\}_{t=1}^{T}$.
Specifically, for each boundary superpoint in $\mathbf{U}_i$, we allow region adjustments based on its affinity scores with adjacent regions, reassigning it to the region with the highest score.
This process iterates until the superpoints are assigned to high-affinity regions, or the predefined stopping criterion is reached.
The proposed refinement step mitigates errors caused by fixed-threshold merging, resulting in more accurate 3D instance segmentation.
The detailed algorithm is provided in the supplementary material.

\section{Experiments}
We evaluate MV3DIS on multiple datasets to validate its effectiveness in zero-shot 3D instance segmentation.
We also compare MV3DIS with leading open-vocabulary segmentation methods and state-of-the-art closed-vocabulary pre-trained segmentation methods.
\subsection{Experiment settings}
\noindent\textbf{Datasets.}
We evaluate our method on three widely used indoor 3D datasets: ScanNetV2 \cite{scannet}, ScanNet200 \cite{scannet200}, and ScanNet++ \cite{scannet++}. 
ScanNetV2 is a large-scale RGB-D dataset comprising 1201 training scenes and 312 validation scenes, covering 18 instance categories.
ScanNet200 extends ScanNetV2 with the same point clouds but richer annotations over 198 categories.
ScanNet++ is a high-quality indoor dataset captured using a laser scanner, DSLR camera, and iPhone RGB-D streams, comprising 460 scenes, 280k DSLR images, and 3.7M RGB-D frames.
To further verify the robustness of our method, we also conduct experiments on the Replica \cite{replica} and Matterport3D \cite{matterport3d} datasets, with results detailed in the supplementary material.

\noindent\textbf{Evaluation metrics.}
We conduct evaluations in two settings: class-agnostic instance segmentation and semantic instance segmentation.
In the class-agnostic setting, category information is ignored, focusing solely on the accuracy of instance masks, with all confidence scores set to 1.0. The semantic setting builds upon this by incorporating semantic labels for each instance.
For both settings, we evaluate the results using the widely adopted Average Precision (AP) score. Following the evaluation scheme from ScanNetV2 \cite{scannet}, we report AP at IoU thresholds of 0.25 (AP$_{25}$) and 0.50 (AP$_{50}$), as well as the average AP over IoU thresholds from 0.50 to 0.95 in steps of 0.05 (mAP).

\noindent\textbf{Implementation details.}
Following recent works~\cite{open3dis,sai3d,openmask3d}, to balance performance and efficiency, we adopt different view sampling rates across datasets. Specifically, we sample 10\% of views for both ScanNetV2 and ScanNet200, and 5\% for ScanNet++.
Additionally, for 2D foundation models, we employ Grounded-SAM (GD-SAM)~\cite{groundingsam} on ScanNetV2 and ScanNet200, while using SAM2~\cite{sam2} for automatic mask generation on ScanNet++ to achieve better granularity control.

\noindent\textbf{Baselines.}
We evaluate our method by comparing it with both closed-vocabulary and open-vocabulary approaches.
For closed-vocabulary methods, we benchmark against the state-of-the-art Mask3D~\cite{mask3d}, which is supervised on training data.
For open-vocabulary methods, we compare with SAM-based approaches, including SAM3D~\cite{sam3d}, SAI3D~\cite{sai3d}, SAM-graph~\cite{samgraph}, Open3DIS~\cite{open3dis}, SAMPro3D~\cite{sampro3d}, Any3DIS~\cite{any3dis}, SGS-3D~\cite{wang2025sgs}, MaskClustering~\cite{maskclustering}, and SAM2Object~\cite{sam2object}.
We also compare with the traditional point grouping method, Felzenszwalb’s algorithm~\cite{superpoint}.
Additionally, in the semantic instance segmentation setting, we compare with OpenMask3D~\cite{openmask3d}, which generates class-agnostic masks using Mask3D and assigns semantic labels with CLIP.

\subsection{Results}
\noindent\textbf{Class-Agnostic 3D Instance Segmentation.}
We first evaluate our method in the class-agnostic 3D instance segmentation setting.
\Cref{tab:scannet-comparison,tab:scannet200-comparison,tab:scannet++-comparison} report quantitative comparisons on the ScanNetV2, ScanNet200, and ScanNet++ datasets, respectively.
Our proposed MV3DIS consistently achieves the best performance among zero-shot open-vocabulary methods across all three benchmarks.
On the ScanNetV2 dataset, MV3DIS significantly outperforms the previous state-of-the-art method SAM2Object~\cite{sam2object}, with gains of 4.5, 7.5, and 5.9 in mAP, AP$_{50}$, and AP$_{25}$, respectively.
On the ScanNet200 dataset, MV3DIS attains 54.7 AP$_{50}$ and 69.7 AP$_{25}$. Notably, this performance even surpasses that of Mask3D~\cite{mask3d} fully trained on ScanNet200 (51.2 AP$_{50}$ and 57.1 AP$_{25}$).
These results highlight the advantages of our method over previous approaches.
On the more challenging ScanNet++ dataset, MV3DIS achieves 22.0 mAP, 36.7 AP$_{50}$, and 51.7 AP$_{25}$, outperforming Open3DIS~\cite{open3dis} and SAM2Object~\cite{sam2object}.
Furthermore, we observe that Mask3D trained on ScanNetV2 performs worse than zero-shot open-vocabulary methods on ScanNet++, indicating that supervised methods lack generalization to more challenging open-vocabulary indoor scenes.

\begin{table}[t]
\centering
\footnotesize
\caption{\textbf{Class-agnostic 3D instance segmentation on ScanNetV2.}
Best and second-best results are bold and underlined, respectively. 
VFM represents the 2D vision foundation model, and Sem-SAM represents Semantic-SAM~\cite{li2023semantic}.}
\label{tab:scannet-comparison}
\setlength{\tabcolsep}{8pt}
\begin{tabular}{lcccc}
\toprule
Method & VFM & mAP & AP$_{50}$ & AP$_{25}$ \\
\midrule
\textit{Closed-vocabulary}\\
Mask3D \cite{mask3d} & None & 65.7 & 83.1 & 91.0 \\
\midrule
\textit{Open-vocabulary}\\
Felzenszwalb \cite{superpoint} &None &5.0 &12.7 &38.9\\
SAM-graph \cite{samgraph} &SAM &24.1  &40.3  &65.9 \\
SAM3D \cite{sam3d} & SAM & 20.2  & 34.0  & 53.3 \\
SAMPro3D\cite{sampro3d}& SAM & 18.0 &32.8 &56.3\\
SAI3D \cite{sai3d} & Sem-SAM & 30.8 & 50.5 & \underline{70.6} \\
SAM2Object \cite{sam2object} & SAM2 & \underline{34.0} & \underline{52.7} & 70.3 \\
SAM3D \cite{sam3d} & GD-SAM & 19.7   & 34.1   & 54.5 \\
Open3DIS \cite{open3dis} & GD-SAM & 31.5 & 48.7 & 59.8 \\
Ours & GD-SAM & \textbf{38.5} & \textbf{60.2} & \textbf{76.2} \\
\bottomrule
\end{tabular}
\end{table}

\begin{table}[t]
\centering
\footnotesize
\caption{\textbf{Class-agnostic 3D instance segmentation on ScanNet200.}
Best and second-best results are bold and underlined, respectively. 
VFM represents the 2D vision foundation model, Sem-SAM represents Semantic-SAM, and CropFormer refers to~\cite{qi2022high}.}
\label{tab:scannet200-comparison}
\setlength{\tabcolsep}{8pt}
\begin{tabular}{lcccc}
\toprule
Method & VFM & mAP & AP$_{50}$ & AP$_{25}$ \\
\midrule
\textit{Closed-vocabulary}\\
Mask3D \cite{mask3d} & None & 40.4 & 51.2 & 57.1 \\
\midrule
\textit{Open-vocabulary}\\
Felzenszwalb \cite{superpoint} &None &4.8 &9.8  &27.5\\
SAM-graph \cite{samgraph} &SAM &22.1  &41.7  &62.8 \\
SAM3D \cite{sam3d} & SAM & 20.0  & 33.8  & 52.7 \\
SAI3D \cite{sai3d} & Sem-SAM & 29.3 & 48.4 & \underline{67.1} \\
SAM3D \cite{sam3d} & GD-SAM & 18.0    & 30.8    & 49.5 \\
Open3DIS \cite{open3dis} & GD-SAM & 29.7 & 45.2 & 56.8 \\
Any3DIS \cite{any3dis} & SAM2 &32.5  &45.2  &55.0 \\
SGS-3D \cite{wang2025sgs} & GD-SAM & \underline{34.3} &\underline{51.3}  &64.6 \\
MaskClustering \cite{maskclustering} & CropFormer &19.7  &36.4  &51.4 \\
Ours & GD-SAM & \textbf{35.5} & \textbf{54.7} & \textbf{69.7} \\
\bottomrule
\end{tabular}
\end{table}

\begin{table}[t]
\centering
\footnotesize
\caption{\textbf{Class-agnostic 3D instance segmentation on ScanNet++.}
Best and second-best results are bold and underlined, respectively. 
VFM represents the 2D vision foundation model, and SAM-HQ refers to \cite{samhq}. Note that Mask3D is trained on the ScanNetV2 dataset.}
\label{tab:scannet++-comparison}
\setlength{\tabcolsep}{8pt}
\begin{tabular}{lcccc}
\toprule
Method & VFM & mAP & AP$_{50}$ & AP$_{25}$ \\
\midrule
\textit{Closed-vocabulary}\\
Mask3D \cite{mask3d} & None & 9.9  & 17.3  & 25.8 \\
\midrule
\textit{Open-vocabulary}\\ 
Felzenszwalb \cite{superpoint} &None &4.1  &9.2  &25.3\\
SAM-graph \cite{samgraph} &SAM &12.9   &25.3  &43.6 \\  
SAM3D \cite{sam3d} & SAM & 7.2 & 14.2 & 29.4 \\
SAMPro3D \cite{sampro3d} & SAM-HQ & 16.5 & 31.2 & 47.6 \\
SAI3D \cite{sai3d} & SAM-HQ & 17.1 & 31.1 & \underline{49.5}  \\
SAM2Object \cite{sam2object} & SAM2 & \underline{20.2} & \underline{34.1} & 48.7 \\
SAM3D \cite{sam3d} & SAM2 &6.9   &13.4   &26.7  \\
Open3DIS \cite{open3dis} & SAM2 & 19.8    & 33.5    & 43.1 \\
Ours & SAM2 & \textbf{22.0} & \textbf{36.7} & \textbf{51.7} \\
\bottomrule
\end{tabular}
\end{table}

\noindent\textbf{Semantic instance segmentation.}
Next, we evaluate our method in the open-vocabulary semantic instance segmentation setting. 
Following previous works \cite{sai3d,sam2object}, we adopt the label assignment strategy from OpenMask3D~\cite{openmask3d} to assign semantic labels to segmented class-agnostic instances. Open3DIS \cite{open3dis} selects the Top-K predictions based on class prediction scores across all instances, enabling multi-label predictions for each 3D instance.
In contrast, other methods \cite{ovir3d,sam3d,sai3d,sam2object,openmask3d} assign single-label predictions to each instance. 
For fair comparison, we evaluate our method under both single-label and multi-label settings. 
As shown in \cref{tab:scannet200ov-comparison}, MV3DIS achieves mAP scores of 15.5 and 20.5 in the two settings, outperforming SAM2Object and Open3DIS by 2.2 and 2.3, respectively. 
Notably, our method attains an AP of 18.9 on tail classes, surpassing OpenMask3D by 4.0. 
This indicates that our method has better generalization ability for rare classes.

\begin{table}[t]
\centering
\footnotesize
\caption{\textbf{Semantic instance segmentation on ScanNet200.}
Head~(AP), Common (AP), and Tail (AP) represent AP scores evaluated on frequent, common, and rare classes, respectively.
``Sup.'' and ``OV.'' represent ``Supervised'' and ``Open-vocabulary''.
Best results are in bold, and the second-best results are underlined.}
\label{tab:scannet200ov-comparison}
\setlength{\tabcolsep}{4pt}
\begin{tabular}{lcccccc}
\toprule
Method  & mAP & AP$_{50}$ & AP$_{25}$ & \makecell{Head \\ (AP)} & \makecell{Common \\ (AP)} & \makecell{Tail \\ (AP)} \\
\midrule
\multicolumn{7}{l}{\textit{Sup. mask + Single-label}} \\
OpenMask3D \cite{openmask3d}  & 15.4 & 19.9 & 23.1 & 17.1 & 14.1 & 14.9 \\
\midrule
\multicolumn{7}{l}{\textit{OV. mask + Single-label}} \\
OVIR-3D \cite{ovir3d}   & 9.3  & 18.7  & \textbf{25.0}  & 9.8  & 9.4  & 8.5 \\
SAM3D \cite{sam3d}  & 9.8 & 15.2 & 20.7 & 9.2  & 8.3  & 12.3 \\
SAI3D \cite{sai3d}  & 12.7 &  18.8 &  24.1 &  \underline{12.1}  & \underline{10.4}  & \underline{16.2} \\
SAM2Object \cite{sam2object}  & \underline{13.3}  &  \underline{19.0}  &  23.8 &  -  & -  & - \\
\textbf{Ours} & \textbf{15.5} & \textbf{20.2} & \underline{24.2} & \textbf{14.2} & \textbf{13.9} & \textbf{18.9} \\
\midrule
\multicolumn{7}{l}{\textit{OV. mask + Multi-label}} \\
Open3DIS \cite{open3dis}    &\underline{18.2}   & \underline{26.1}   & \underline{31.4}   & \underline{18.9} & \underline{16.5} & \underline{19.2}  \\
\textbf{Ours}  & \textbf{20.5} & \textbf{27.7} & \textbf{33.4} & \textbf{20.3} & \textbf{18.5} & \textbf{23.1} \\
\bottomrule
\end{tabular}
\end{table}

\begin{figure*}[t]
    \centering
    \includegraphics[width=0.85\textwidth]{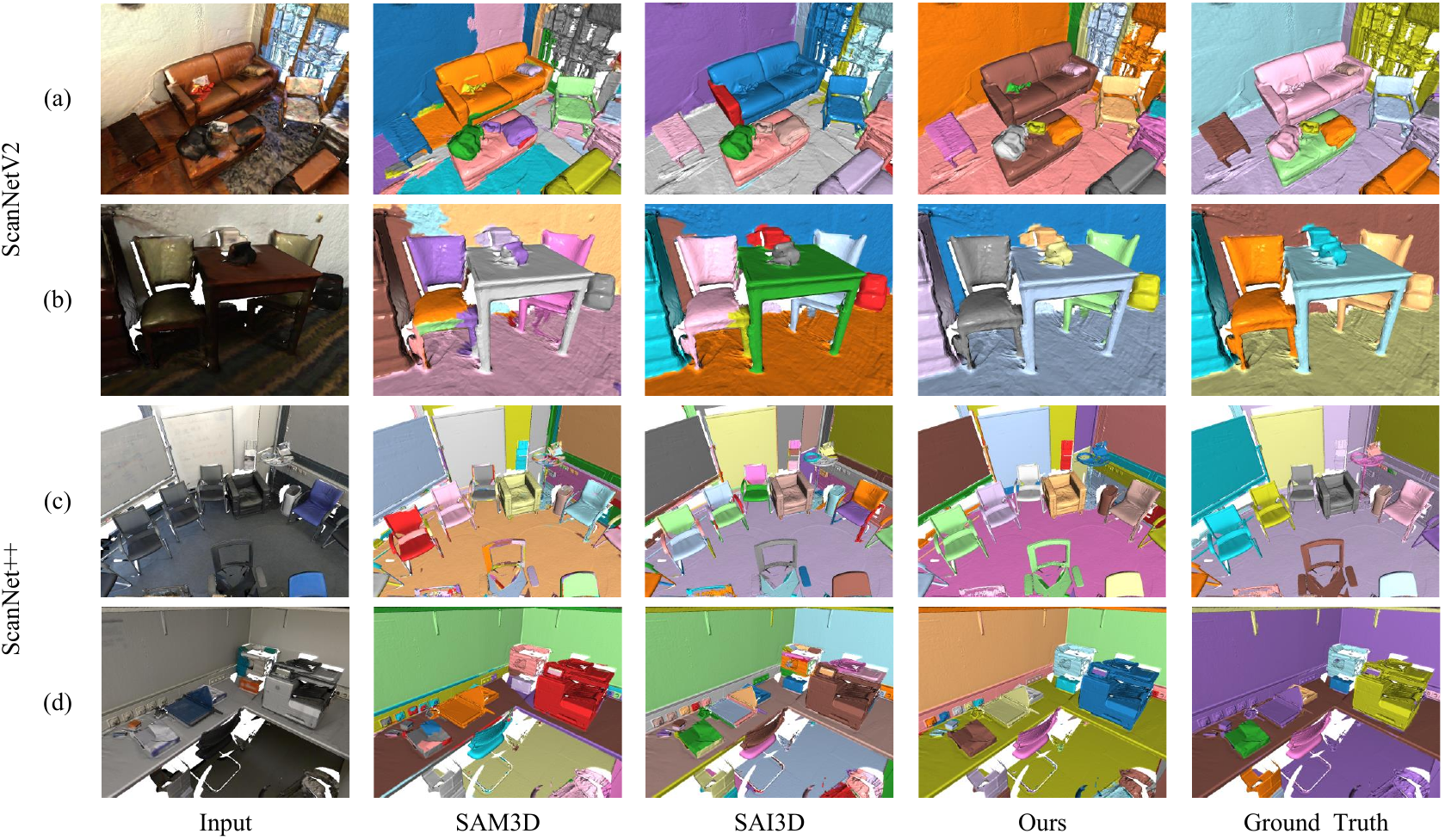}
    \caption{Qualitative results of our method compared to SAM3D \cite{sam3d} and SAI3D \cite{sai3d} on the ScanNetV2 and ScanNet++ datasets.}
    \label{Visio-vis_v2_++}
\end{figure*}

\noindent\textbf{Qualitative results.}
In \cref{Visio-vis_v2_++}, we present qualitative results on the ScanNetV2 and ScanNet++ datasets. 
Due to a lack of view consistency, baseline methods such as SAM3D and SAI3D exhibit over-segmentation on the sofa in \cref{Visio-vis_v2_++} (a) and the chairs in Figs.~\ref{Visio-vis_v2_++}~(b) and (c). 
In contrast, our method achieves high-quality 3D instance segmentation by enhancing multi-view consistency.  
Furthermore, leveraging SAM's fine-grained segmentation capability, our method produces masks with finer granularity than the ground truth annotations.
As shown in \cref{Visio-vis_v2_++}~(d), it separates the network ports and power outlets from the wall, whereas they are merged into the wall region in the ground truth.

\subsection{Ablation and Analysis}
\noindent\textbf{Study on the effect of each component.}
\Cref{tab:ablation-main-simplified} details the contributions of each component. 
We analyze the three key designs of MV3DIS: region refinement (RR), 3D-guided mask matching (3DG-MM), and depth consistency weight (DCW).
We start from a baseline that relies solely on coarse 3D segmentation, yielding 33.3 mAP.
When adding region refinement, the performance improves from 33.3 mAP to 35.5 mAP, highlighting its role in refining superpoint assignments.
Incorporating 3DG-MM further increases the mAP to 37.0.
This gain confirms that enhancing multi-view mask consistency facilitates more precise affinity calculations, resulting in more accurate 3D segmentation.
Finally, including the depth consistency weight yields 38.5 mAP, confirming its effectiveness in assessing projection reliability.
Overall, our full method achieves a gain of 5.2 mAP over the baseline, demonstrating the effectiveness of the proposed components.
\begin{table}[t]
\centering
\footnotesize
\caption{\textbf{Ablation study of our main components.}
RR indicates region refinement, 
3DG-MM indicates 3D-guided mask matching, 
and DCW indicates depth consistency weight.}
\label{tab:ablation-main-simplified}
\setlength{\tabcolsep}{9pt}
\begin{tabular}{cccccc} 
\toprule
RR & 3DG-MM & DCW & mAP & AP$_{50}$ & AP$_{25}$ \\
\midrule
 &  &  & 33.3 & 51.3 & 70.1 \\
$\checkmark$ &  & & 35.5 & 53.2 & 72.1 \\
$\checkmark$ & $\checkmark$ &  & 37.0 & 57.6 & 74.2 \\
$\checkmark$ & $\checkmark$ & $\checkmark$ & \textbf{38.5} & \textbf{60.2} & \textbf{76.2} \\
\bottomrule
\end{tabular}
\end{table}

\noindent\textbf{Hyperparameter analysis.}
We conduct a sensitivity analysis on the ScanNetV2 dataset for two key hyperparameters in MV3DIS: the depth tolerance $\alpha$ (\cref{tab:ablation-tau}) and the merging threshold $\tau_{\mathrm{merge}}$ (\cref{tab:ablation-merge}) used in region growing.
Both hyperparameters exhibit a similar trend, where mAP first increases and then decreases as the value grows.
For $\alpha$, a small value imposes an overly strict criterion for depth consistency, discarding numerous valid projected points and underutilizing 2D masks, while a large value sets an overly loose criterion, introducing occluded points that adversely affect the affinity estimation.
For $\tau_{\mathrm{merge}}$, low values relax the merging criterion, leading to over-merging and under-segmentation, whereas high values impose a strict criterion, causing insufficient merging and over-segmentation.
\begin{table}[t]
  \centering
\footnotesize
  \hspace{0.005\linewidth}
  \begin{minipage}[b]{0.46\columnwidth}
    \caption{Ablation on $\alpha$.}
    \label{tab:ablation-tau}
    \setlength{\tabcolsep}{3pt}
    \begin{tabular*}{\linewidth}{@{\extracolsep{\fill}}c S[table-format=2.1] S[table-format=2.1] S[table-format=2.1]}
      \toprule
      $\alpha$ & {mAP} & AP$_{50}$ & AP$_{25}$ \\
      \midrule
      0.01 & 37.4 & 59.0 & 75.7 \\
      0.05 & \textbf{38.5} & \textbf{60.2} & 76.2 \\
      0.10 & 37.9 & 60.2 & \textbf{76.7} \\
      0.20 & 36.0 & 58.6 & 75.8 \\
      \bottomrule
    \end{tabular*}
  \end{minipage}
  \hfill
  \begin{minipage}[b]{0.46\columnwidth}
    \caption{Ablation on $\tau_{\mathrm{merge}}$.}
    \label{tab:ablation-merge}
    \setlength{\tabcolsep}{3pt}
    \begin{tabular*}{\linewidth}{@{\extracolsep{\fill}}c S[table-format=2.1] S[table-format=2.1] S[table-format=2.1]}
      \toprule
      $\tau_{\mathrm{merge}}$ & {mAP} & AP$_{50}$ & AP$_{25}$ \\
      \midrule
      0.3 & 33.4  & 53.1 & 70.3 \\                     
      0.5 & \textbf{38.5} & \textbf{60.2} & \textbf{76.2} \\
      0.6 & 38.0 & 58.9 & 75.8 \\
      0.9 & 31.7  & 50.3  &69.9   \\
      \bottomrule
    \end{tabular*}
  \end{minipage}
  \hspace{0.005\linewidth}
\end{table}
\begin{figure}[t!]
    \centering
    \includegraphics[width=0.9\columnwidth]{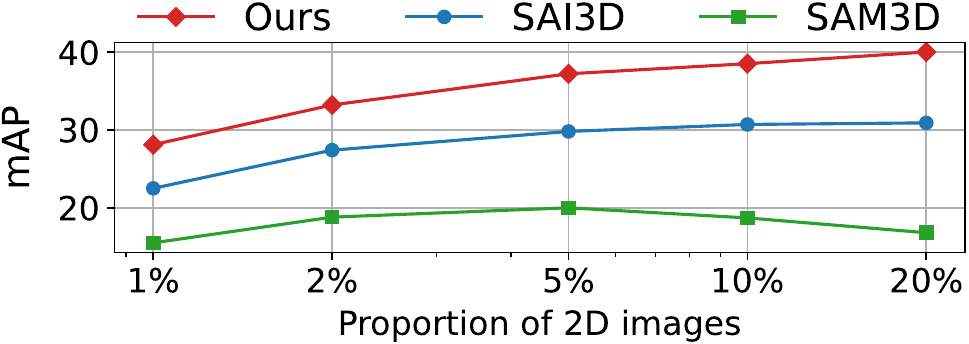}
    \caption{Correlation of performance with the number of images.}
    \label{Figure_1}
\end{figure}

\noindent\textbf{Effect of the number of 2D images.}
We evaluate the performance of MV3DIS under different proportions of input views on the ScanNetV2 dataset.
As shown in~\cref{Figure_1}, as the sampled view ratio increases from 1\% to 20\%, our method's performance steadily improves and consistently outperforms SAI3D and SAM3D.
In contrast, SAM3D, which employs an adjacent frame merging strategy, struggles to effectively aggregate information beyond a 5\% proportion.
Furthermore, our method using only 2\% of images surpasses SAI3D using 20\% of images, highlighting its stability under sparse view sampling.

\section{Conclusion}
In this paper, we propose MV3DIS, a zero-shot 3D instance segmentation framework.
The core of our approach is a 3D-guided mask matching strategy that effectively mitigates view inconsistencies arising when lifting segmentations from 2D foundation models to 3D scenes.
Specifically, we leverage coarse 3D segment projections as a common reference to associate 2D masks and then quantify their consistency according to the 3D coverage distribution.
Subsequently, these consistent masks guide the region refinement process, merging coarse 3D segments into coherent 3D instances.
Extensive experiments on datasets such as ScanNetV2, ScanNet200, and ScanNet++ demonstrate that MV3DIS significantly outperforms previous methods in both class-agnostic and open-vocabulary settings.

\section*{Acknowledgements}
This work was supported by the National Science Fund of China under Grant Nos. 62361166670, U62276144 and 62306155.

{
    \small
    \bibliographystyle{ieeenat_fullname}
    \bibliography{main}
}


\end{document}